\pdfoutput=1

\documentclass[11pt]{article}

\usepackage[]{ACL2023}

\usepackage{times}
\usepackage{latexsym}

\usepackage{algorithm}
\usepackage[noend]{algorithmic}
\usepackage[center]{subfigure}
\usepackage{color,graphicx}
\usepackage{mathtools}
\usepackage{amsmath}
\usepackage{amssymb}
\usepackage{url}
\usepackage{multicol}
\usepackage{multirow}
\usepackage{booktabs}
\usepackage{pifont}
\newcommand{\cmark}{\ding{51}}%
\newcommand{\xmark}{\ding{55}}%

\usepackage[T1]{fontenc}

\usepackage[utf8]{inputenc}

\usepackage{microtype}

\usepackage{inconsolata}

\title{Are LLMs Rigorous Logical Reasoners? Empowering Natural Language Proof Generation by  Stepwise Decoding with Contrastive Learning}

\author{Ying Su$^1$,  Mingwen Liu$^2$, Zhijiang Guo$^{3,4}$\\
  South China University of Technology$^1$, Likelihood Lab$^2$, HKUST(GZ)$^3$, HKUST$^4$\\
  \texttt{yingsu@scut.edu.cn}, 
  \texttt{maxwell@alphafuture.cn}, \texttt{zhijiangguo@hkust-gz.edu.cn} \\}

\begin{document}
\maketitle


\begin{abstract}
Logical reasoning is a pivotal component in the field of artificial intelligence. 
Proof planning, particularly in contexts requiring the validation of explanation accuracy, continues to present challenges. 
The recent advancement of large language models (LLMs) has led to significant progress in natural language proof planning, evolving from one-stage generators to more complex three-stage systems that include additional searchers or verifiers. 
While these assisted methods improve the quality of generated results, they also introduce increased search efforts and computational costs. 
Furthermore, the generative process itself remains underexplored.
In this study, we propose a stepwise decoding approach augmented by contrastive learning to address two common errors encountered during the LLM generator's decoding process. 
We fine-tune the language model using both vanilla and enhanced hard negatives to mitigate these decoding errors.
Empirical results demonstrate the effectiveness of our strategy. 
Additionally, our further analysis reveals that even larger LLMs still struggle to generate rigorous logical chains.
\end{abstract}
\section{Introduction}

Logical reasoning underpins the comprehension of human cognition and intelligence in machines~\citep{goel2017reasoning}. Large Language Models (LLMs) like GPT~\citep{brown2020language,ouyang2022training} and PaLM~\citep{chowdhery2022palm,anil2023palm} have pioneered using natural language as a platform for logical reasoning, complementing the traditional use of formal languages~\citep{kazemi2022lambada,creswell2022selection}. The incorporation of natural language broadens the scope of logical reasoning by allowing flexible querying and tapping into the extensive implicit knowledge encapsulated within LLMs.

In examining the capacity of LLMs for logical reasoning, it is crucial to consider not only the accuracy of their answers but also the correctness of their explanations~\citep{xu2023large}. Utilizing prompting methods such as in-context learning~\citep{brown2020language} and chain-of-thought~\citep{WeiCoT22}, LLMs have shown promising results across various deductive reasoning tasks in question-answering formats \cite{weston2015towards, tafjord2021proofwriter, saparov2022language, han2022folio}. These approaches decompose the final task goal by guiding the LLMs through intermediate reasoning steps in a carefully constructed context. However, providing correct explanations, which covers \textit{completeness}, \textit{redundancy}, \textit{correctness}~\citep{xu2023large}, emerges as a more daunting challenge. This is particularly evident in tasks that involve generating reasoning chains from premises leading to a conclusion, known as proof generation~\cite{ClarkTR20, dalvi2021explaining}. Unfortunately, LLMs often fall short in creating concise and exact proof trees, commonly producing superfluous or imprecise intermediate steps.

\begin{table*}[t]
\centering
\scalebox{0.8}{\begin{tabular}{p{5.7cm}p{1.9cm}p{1.7cm}p{1.3cm}p{1.0cm}p{1.0cm}p{2.8cm}p{1.0cm}}
 \toprule
 Method & Stepwise Generation & Stepwise Correction & Direction & Search & Verifier & Human-authored Benchmark & Stage \\
 \midrule
 EntailmentWriter~\citep{dalvi2021explaining}  & \textcolor{orange}{\xmark} & \textcolor{orange}{\xmark} & $\rightarrow$ & \textcolor{orange}{\xmark} & \textcolor{orange}{\xmark} & \textcolor{blue}{\cmark} & 1 \\
 IRGR~\citep{ribeiro2022entailment}  & \textcolor{blue}{\cmark} & \textcolor{orange}{\xmark} & $\rightarrow$ & \textcolor{blue}{\cmark}  & \textcolor{orange}{\xmark} & \textcolor{blue}{\cmark} & 2 \\
 SCSearch~\citep{bostrom2022natural}  & \textcolor{blue}{\cmark} & \textcolor{orange}{\xmark} & $\rightarrow$ & \textcolor{blue}{\cmark} & \textcolor{orange}{\xmark} & \textcolor{orange}{\xmark} & 2 \\
 MetGen~\citep{hong-etal-2022-metgen}  & \textcolor{blue}{\cmark} & \textcolor{orange}{\xmark} & both & \textcolor{blue}{\cmark} & \textcolor{orange}{\xmark} & \textcolor{blue}{\cmark} & 2 \\
 ADGV~\citep{sprague2022natural} & \textcolor{blue}{\cmark} & \textcolor{orange}{\xmark} & both & \textcolor{blue}{\cmark} & \textcolor{blue}{\cmark} & \textcolor{blue}{\cmark} & 3 \\
 NLProofs~\citep{yang2022generating} & \textcolor{blue}{\cmark} & \textcolor{orange}{\xmark} & $\rightarrow$ & \textcolor{blue}{\cmark} & \textcolor{blue}{\cmark} & \textcolor{blue}{\cmark} & 3 \\
 \midrule
 ConDec & \textcolor{blue}{\cmark} & \textcolor{blue}{\cmark} & $\rightarrow$ & \textcolor{orange}{\xmark} & \textcolor{orange}{\xmark} & \textcolor{blue}{\cmark} & 1 \\
 \bottomrule
\end{tabular}}
\caption{Comparison of methods over natural language proof generation. \textit{Stepwise Correction} means that if stepwise generation is enhanced in training. \textit{Stage} calculates if the method contains generation, verification, and search.}
\label{tab:1}
\vskip-0.5em
\end{table*}



Previous studies have utilized LLMs to generate proof trees, employing a range of techniques from holistic approaches~\citep{qu-etal-2022-interpretable} to incremental steps~\citep{tafjord2021proofwriter, sanyal-etal-2022-fairr}. Recent methods increasingly rely on post-processing to enhance the quality of generated results, introducing verification- and search-based systems~\citep{hong-etal-2022-metgen, yang2022generating}. However, as the methods become more complex, there is a corresponding increase in search efforts and computational costs. Conversely, there has been insufficient focus on refining the generative process itself. Current models exhibit proficiency in selecting relevant premises but struggle to deduce intermediary conclusions, highlighting a deficiency in their understanding of semantic nuances during stepwise deductive reasoning.

Addressing this issue, we introduce a novel strategy dubbed ConDec (\textbf{Con}trastive learning based stepwise \textbf{Dec}oding), designed to enhance the generative aspect of LLMs for deductive reasoning tasks. ConDec leverages carefully constructed hard negatives -- outputs that are deceivingly similar in form yet differ semantically -- to refine generation precision. These hard negatives can be simple sequence alterations or products of an intricate sampling and reasoning process, aided by an external reasoner and checker. Intuitively, the hard negatives are designed to solve decoding errors analyzed in \cite{dalvi2021explaining}: \textit{repetition} and \textit{invalid entailment}. Finetuning with these hard negatives notably advances the LLMs' proficiency in intermediate step and conclusion generation, culminating in overall improved proof accuracy. The main contributions of this study are threefold:
\begin{itemize}
    \item We introduce ConDec, a stepwise decoding with contrastive learning strategy that enhances stepwise generative quality in proof generation tasks, and devise an automatic method for hard negative generation involving a reasoner and a checker;
    \item We conduct an extensive empirical analysis on the Entailment benchmark, demonstrating the effectiveness of the proposed method;
    \item We reveal that LLMs even equipped with chain-of-thought strategies still struggle to perform rigorous logical reasoning in natural language proof generation tasks.
\end{itemize}
\section{Related Work}

\subsection{Logical Reasoning with Natural Language}
Logical reasoning is an important ability to realize human-level cognition and intelligence in AI~ \citep{Nunes2012logic}. Early research of logical reasoning uses formal language to represent knowledge and conducts symbolic reasoning~\citep{muggleton1994inductive}. Recent research uses pretrained language models for logical reasoning in the form of natural language to alleviate the representation challenge with formal language~\citep{musen1988brittleness}. 

Among the logical reasoning over natural language \cite{yang2023logical}, deductive reasoning covers aspects including hypothesis classification, proof generation, proof generation with incomplete information, and implication enumeration. Several tasks have been proposed to evaluate these reasoning abilities. Specifically, hypothesis classification is conducted over RuleTaker with transformers~\citep{ClarkTR20}. Proof generation providing rationals along with the predicted answer for emulating formal reasoning is further proposed to increase the explanability~\citep{saha2020prover}. ProofWriter~\citep{tafjord2021proofwriter} produces a deductive chain of reasoning over proof generation and implication enumeration with an iterative generating style. To enhance the chain of reasoning for multi-step premises, EntailmentWriter tests transformers' explainability in the form of entailment trees over EntailmentBank~\citep{dalvi2021explaining}. 

\begin{figure*}[t]
\centering
\includegraphics[scale=0.47,trim={0cm 0cm 0cm 0cm}]{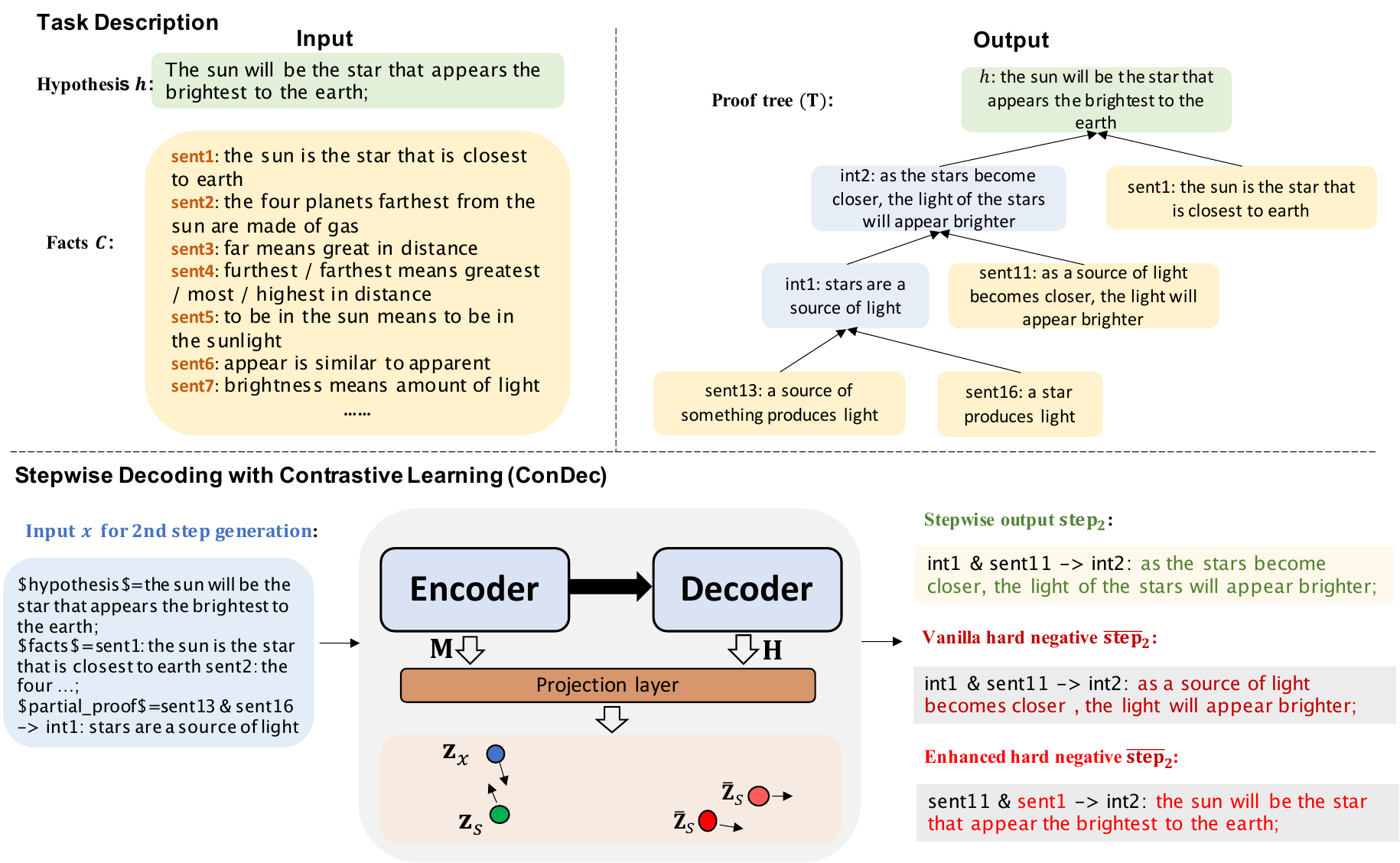}
\caption{Architecture of the stepwise decoding with contrastive learning over hard negatives. The hard negatives are constructed by vanilla and enhanced strategies. Vanilla strategy means simple conclusion substitution. The enhanced strategy uses a reasoner and a checker to generate hard negatives.}
\vskip-0.5em
\label{fig:interactive}
\end{figure*}

\subsection{Proof Generation}
Methods for finetuning language models for proof generation vary in the proof direction, inference with or without hypothesis, and whether search or verification is involved. One line of research is inference without a hypothesis available. FaiRR~\citep{sanyal-etal-2022-fairr} breaks proof generation into three steps: rule selection, fact selection, and knowledge composition. MetGen~\citep{hong-etal-2022-metgen} iteratively generates the entailment tree by conducting a single-step entailment with separate modules and a reasoning controller. SCSearch~\citep{bostrom2022natural} decomposes the deductive reasoning task into separate steps coordinated by a search procedure, producing a tree of intermediate conclusions that faithfully reflects the system's reasoning process. ADGV~\citep{sprague2022natural} proposes to abductively infer a premise given another premise and a conclusion, as well as to search over two fingers interleaving deductive (forward-chaining) and abductive (backward-chaining) inferences. 

Another line of work is with the hypothesis available. IBR~\citep{qu-etal-2022-interpretable} enhances the interpretability of reasoning procedures by predicting nodes and edges in the proof tree iteratively backward from the question, as well as increasing efficiency and accuracy by simplifying the intermediate process of reasoning. IRGR~\citep{ribeiro2022entailment} explains a given hypothesis by iteratively searching for suitable premises, constructing a single entailment step at a time. 
NLProofs~\citep{yang2022generating} trains an independent verifier to check the validity of the proof steps to improve decoding accuracy. 
Our work follows this line and we focus on stepwise decoding correction on the generator itself. A comparison of our method with other approaches with LLMs is presented in Table \ref{tab:1}.


\section{Problem Formulation}
Following the task definition in \citet{yang2022generating}, the proof generation task is to derive a proof tree $T$ given a hypothesis $h$ and a set of supporting facts $C=\{{\rm sent}_1, {\rm sent}_2, ..., {\rm sent}_n$\}. The proof tree $T$ is represented as a tuple $T=(h, \hat{C}, U, S)$. $\hat{C}$ is a subset of the facts $\{{\rm sent}_i\} \in C$, denoting leaf nodes on the tree $T$. $U=\{{\rm int}_1, {\rm int}_2, ..., {\rm int}_m\}$ denotes the intermediate nodes on the tree. The intermediate nodes are deduced during the reasoning process. Each intermediate node represents an intermediate conclusion. The intermediate nodes are internal tree nodes. On top of the tree, $h$ is the root node as well as the final conclusion needs to be proved. 

The structure of the tree is denoted by reasoning steps $S=\{{\rm step}_1, {\rm step}_2, ...,{\rm step}_t\}$. Each internal tree node corresponds to a reasoning step ${\rm step}_i \in S$ with ${\rm int}_j \in U$ as the conclusion and its children as premises, i.e., ${\rm sent}_1$ \& ${\rm int}_1 \rightarrow {\rm int}_2$, representing that intermediate node ${\rm int}_2$ is the conclusion of leaf node ${\rm sent}_1$ and intermediate node ${\rm int}_1$. The premise of a reasoning step can be from leaf nodes or intermediate nodes. Under the stepwise generation setting, intermediate nodes and proof steps up to the current step are added to given facts as input to generate new intermediate nodes for the next step. 

\section{Approach}

\subsection{Stepwise Decoding with Contrastive Learning}
With stepwise training, subtrees are sampled on the original entire proof graph. The decoding goal can be the intermediate node or the final hypothesis of the subtree, depending on the sampling strategy. As analyzed in \cite{dalvi2021explaining}, there are decoding errors leading to inaccurate proof step generation, finally leading to entire proof generation failure. An overview is presented in Figure \ref{fig:interactive}.

To address the problem, we adopt a contrastive learning technique to improve the stepwise decoding quality. Learning with contrasting positive and negative pairs can improve the generalization ability of conditional text generations \cite{lee2020contrastive, an2022cont}. Inspired by this, we construct negative decoding samples to improve the reasoning ability of the generator.

The goal of the generator is to output a stepwise reasoning step ${\rm step}^{(i)}_j$ with tokens $(\tilde{s}^{(i)}_1, \tilde{s}^{(i)}_2, ..., \tilde{s}^{(i)}_L)$ with length $L$ conditioned on the input text sequence $x^{(i)} = (\tilde{x}^{(i)}_1, \tilde{x}^{(i)}_2,...)$. The input text sequence is a concatenation of contexts from hypothesis $h$, facts $C$, and previous steps $\{{\rm step}^{(i)}_1, ..., {\rm step}^{(i)}_{j-1}\}$. $i$ is the index of instances in a batch. The finetuning loss is to maximize the conditional log-likelihood $\log p_{\theta}({\rm step}_j|x)$ for a given $N$ observations $\{{(x^{(i)}, {\rm step}^{(i)}_j)}^N_{i=1}\}$ as follows:
\begin{equation}
    \mathcal{L}_{MLE}(\theta) = \sum^{N}_{i=1}\log 
 p_{\theta}({\rm step}^{(i)}_j|x^{(i)}), 
\end{equation}
\begin{equation}
    p_{\theta}(\tilde{s}^{(i)}_1, ..., \tilde{s}^{(i)}_L|x^{(i)}) = \prod^{L}_{l=1}p_{\theta}(\tilde{s}^{(i)}_l|\tilde{s}^{(i)}_{<l},x^{(i)}),
\end{equation}

\begin{equation}
    \mathbf{h}^{(i)}_l = g(\tilde{s}^{(i)}_{l-1},\mathbf{M}^{(i)};\theta), \mathbf{M}^{(i)}=f(x^{(i)};\theta),
\end{equation}
where $f,g$ denote the encoder and the decoder respectively. $\mathbf{M}^{(i)}$ is the concatenation of the hidden representations of the source tokens $x^{(i)}$. $\mathbf{H}^{(i)}$ is the concatenation of the hidden states $[\mathbf{h}^{(i)}_1, ..., \mathbf{h}^{(i)}_L]$ at the decoder output.

With a linear projection layer, the hidden states $\mathbf{M}^{(i)}$ and $\mathbf{H}^{(i)}$ of the encoder and decoder are mapped onto the latent embedding space:
\begin{equation}
    \mathbf{z}^{(i)}_{x} = {\rm AvgPool}({\rm ReLU}(\mathbf{W}_{proj}\mathbf{M}^{(i)} + \mathbf{b}_{proj})),
\end{equation}
\begin{equation}
    \mathbf{z}^{(i)}_{s} = {\rm AvgPool}({\rm ReLU}(\mathbf{W}_{proj}\mathbf{H}^{(i)} + \mathbf{b}_{proj})).
\end{equation}
The semantic similarity $sim$ between them can be calculated by distance with a dot or cosine function. A contrastive loss maximizes the similarity between the pair of source sequence and target sequence while minimizing the similarity between the negative pairs as follows:
\begin{equation}
    \mathcal{L}_{cont}(\theta) = \sum^{N}_{i=1}\log \frac{\exp(sim(\mathbf{z}^{(i)}_{x}, \mathbf{z}^{(i)}_{s})/\tau)}{\sum_{\mathbf{z}^{(k)}_{s} \in \mathcal{S}} \exp(sim(\mathbf{z}^{(i)}_{x}, \mathbf{z}^{(k)}_{s})/\tau)} ,
\end{equation}
where $\mathcal{S}$ is the set of negative samples in the same batch and $\tau$ is the temperature parameter.

\begin{figure*}[t]
\centering
\includegraphics[scale=0.55,trim={0cm 0cm 0cm 0.9cm}]{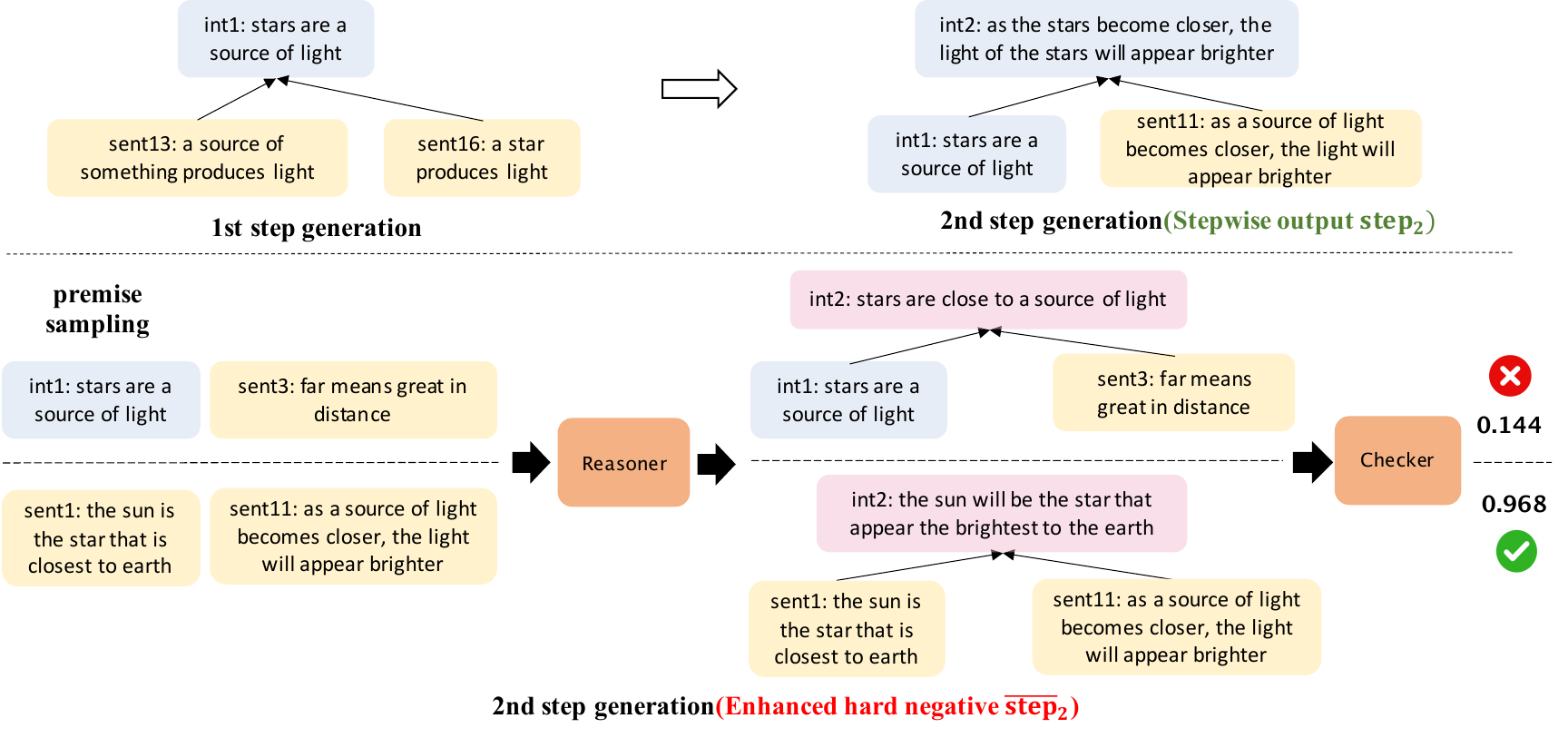}
\caption{Enhanced hard negative construction is implemented by exploring the unseen combination of premises, inferencing with the reasoner, and filtering with a $score$ from the checker.}
\vskip-1.0em
\label{fig:verifier}
\end{figure*}

\subsection{Training with Hard Negative}
Though stepwise generation improves over a single-shot generation (training over the entire proof tree and the decoding goal is a hypothesis), it still has typical errors \cite{dalvi2021explaining}: 1) \textbf{repetition}(the entailed conclusion simply repeats one of the input sentences); 2) \textbf{invalid entailment}(the entailed conclusion does not follow from input sentences). To improve the decoding quality, we design two types of hard negatives for these errors and finetune the model with these hard negatives. 

The hard negative sequence ${\overline{\rm {step}}}^{(i)}_j$ is constructed based on the gold proof step sequence ${\rm {step}}^{(i)}_j$. With the hard negative sequences, the decoding loss becomes:

\begin{equation}
\begin{aligned}
    &\mathcal{L}_{cont-hard}(\theta) = \\
    &\sum^{N}_{i=1}\log \frac{\exp(sim(\mathbf{z}^{(i)}_{x}, \mathbf{z}^{(i)}_{s})/\tau)}{\sum_{\mathbf{z}^{(k)}_{s} \in \mathcal{S} \bigcup \{\mathbf{\bar{z}}^{(i)}_{s}\}}\exp(sim(\mathbf{z}^{(i)}_{x}, \mathbf{z}^{(i)}_{s})/\tau)},
\end{aligned}
\end{equation}
where $\mathbf{\bar{z}}^{(i)}_{s}$ is the projected hidden state of hard negatives. The final loss for finetuning is:
\begin{equation}
    \mathcal{L} = \mathcal{L}_{MLE} + \alpha \mathcal{L}_{cont-hard},
\end{equation}
$\alpha$ is a weighted parameter.

\subsubsection{Vanilla Hard Negative}
Vanilla hard negatives are constructed based on substitution, which mimics the form of gold stepwise proofs, to address the repetition error. we randomly select one of the premises in a proof step ${\rm {step}_j}$ and replace the conclusion with its context. For example in Figure \ref{fig:interactive}, the vanilla hard negative is constructed by replacing the gold standard conclusion with the context of input node \textit{sent11}.   

\subsubsection{Enhanced Hard Negative}
To increase the entailment quality over stepwise proofs, we also propose to construct enhanced hard negatives by exploring unseen proof steps with a reasoner and a checker. The reasoner is first trained with proof steps ${\rm {step}_j} \in S$ in natural language and then utilized to generate a conclusion given an unseen combination of premises in the supporting facts $C$.

Given proof step ${\rm {step}_j}$, the premises and conclusion in natural language are denoted as set $\{p_1, p_2,...\}$ and $c$. The premises are concatenated as input and the conclusion is output for training the reasoner. 
\begin{equation}
    \mathcal{L}_{MLE}(\phi) = \log p_{\phi}(c|p_1, p_2, ...),
\end{equation}


\noindent After training, the reasoner can generate a conclusion given an unseen combination of premises. The premise combinations are first sampled from supporting fact set. The sampled premises $\{p^{s}_1, p^{s}_2, ...\}$ and generated conclusion $c^{s}$ constructs an enhanced hard negative step. Details are presented in Figure \ref{fig:verifier}. In the example, for premises \textit{int1} and \textit{sent11} in gold proof, randomly sample one premise to substitute one of them and use the reasoner to generate a new conclusion given the recombined premises. 

To improve the quality of the hard negatives generated from the reasoner, we further adopt the checker Vera \cite{liu2023vera} to score the hard negatives and filter those with low scores. Vera is finetuned over T5-11B with commonsense datasets. The $score$ from the checker indicates the extent of the reasonableness of the deductive commonsense knowledge generated by the reasoner:
\begin{equation}
    score = {\rm sigmoid}([p^{s}_1, p^{s}_2, ..., c^{s}];\gamma),
\end{equation}
where the $score$ is calculated with a ${\rm sigmoid}$ function after hidden layers $\gamma$.

\section{Experiment}
\subsection{Dataset}

\noindent \textbf{EntailmentBank}~\citep{dalvi2021explaining}: Entailment trees are made up of individual and multi-premise textual entailment steps. EntailmentBank contains 1,840 multi-step entailment trees for accompanying QA pairs. For the proof generation task, only the hypothesis $H$ and context set $C$ are used as inputs. Each proof tree $T$ contains an average of 6.6 nodes and 2.7 entailment steps. Train/Validation/Test splits are 1,313/187/340 respectively. EntailmentBank consists of three tasks as follows:\\
(1) \textbf{Task 1 (no-distractor)}: $C$ consists of exactly the leaf nodes of the ground truth proof tree; \\
(2) \textbf{Task 2 (distractor)}: $C$ consists of 15-20 distractor sentences besides the leaf nodes on the ground truth proof tree; \\
(3) \textbf{Task 3 (full-corpus)}: $C$ is a large corpus of 12K sentences derived from WorldTree V2 \cite{xie2020worldtree}, requiring the model to retrieve relevant supporting facts from the corpus. For each hypothesis, 25 supporting facts are retrieved. Following~\citet{dalvi2021explaining}, we evaluate the zero-shot performance of the model from Task 2. 


\subsection{Evaluation}
\label{sec:appendix 4}
Following~\citet{dalvi2021explaining}, we use the official tools\footnote{https://github.com/allenai/entailment\_bank} to evaluate the generated entailment tree $T=(h, \mathcal{L}, \mathcal{E}, \mathcal{S})$ with the golden entailment tree $T^*=(h, \mathcal{L}^*, \mathcal{E}^*, \mathcal{S}^*$). These metrics evaluate the correctness along 4 dimensions: \\
(1) \textbf{Leaves} (F1, AllCorrect): F1 measures the precision of leaf nodes of $T$ comparing to gold tree $T^*$. ALLCorrect=1 if F1=1, and ALLCorrect=1 if F1$<$1. \\
(2) \textbf{Steps} (F1, AllCorrect): F1 measures the precision of proof steps structurally correctness. Each step contains an internal node $u \in T^*$ (aligned to $v \in T$). The predicted step is correct if $u$ and $v$ are perfectly aligned. For each tree, ALLCorrect=1 if F1=1 otherwise 0. \\
(3) \textbf{Intermediates} (F1, AllCorrect): For the internal node $u \in T^*$ (aligned to $v \in T)$, the intermediate conclusion is correct if the BLEURT \cite{sellam2020bleurt} score between $u$ and $v$ is greater than 0.28 \cite{dalvi2021explaining}. F1 and AllCorrect from all intermediate conclusions in $T^*$ and $T$ are calculated. ALLCorrect=1 if F1=1 otherwise 0. \\
(4) \textbf{Overall} (AllCorrect): The metric evaluates whether leaves, steps, and intermediates are all correct, AllCorret = 1 if and only if all the leaves, steps, and intermediates are all correct. 

\begin{table}[t]
    \centering
    \scalebox{0.9}{
    \begin{tabular}{ccccc}
    \toprule
    Reasoner & \multicolumn{2}{l}{Enhanced Negative} & \multicolumn{2}{l}{Filtered Negative} \\ \hline
\multirow{2}{*}{8,819} &   Task 1   &   Task 2  &   Task 1  &   Task 2 \\ 
      \cline{2-3}  \cline{4-5}           
        &  47,386  &  104,665 &   16,371   & 21,948   \\ 
    \bottomrule
    \end{tabular}}
    \caption{Distribution of samples from training data, constructed enhanced hard negatives and filtered hard negatives on Task 1 and Task 2. In filtering, threshold score is 0.9.}
    \label{tab:data 1}
    \vskip-1.0em
\end{table}

\subsection{Implementation Details}
The optimizer is set as Adam \cite{kingma2014adam} for all the training. The average running time is 18 hours for Task 1 and 24 hours for Task 2. Experiments are conducted on A800. \\

\noindent \textbf{Generator}. We use Flan-T5-Large as the generator. Flan-T5-Large is a finetuned version of T5-Large~\citep{raffel2020exploring} over a collection of FLAN instructions \cite{chung2022scaling, longpre2023flan}. The generator is trained for 500 epochs on Task1 and 600 epochs on Task2. The learning rate for the first-stage generator is 1e-4 and 5e-5 for Task 1 and Task 2 respectively. \\

\begin{table*}[t]
  \centering
  \scalebox{0.8}{
  \begin{tabular}{ccccccccc}
  \toprule
  \multicolumn{1}{c}{\multirow{2}{*}{Task}} & \multicolumn{1}{c}{\multirow{2}{*}{Method}} & \multicolumn{2}{c}{Leaves}                              & \multicolumn{2}{c}{Steps}                               & \multicolumn{2}{c}{Intermediates}                       & \multicolumn{1}{c}{Overall}    \\
  \multicolumn{1}{c}{}                      & \multicolumn{1}{c}{}                        & \multicolumn{1}{c}{F1} & \multicolumn{1}{c}{AllCorrect} & \multicolumn{1}{c}{F1} & \multicolumn{1}{c}{AllCorrect} & \multicolumn{1}{c}{F1} & \multicolumn{1}{c}{AllCorrect} & \multicolumn{1}{c}{AllCorrect} \\
     \midrule
     Task 1 &  EntailmentWriter & 98.7 & 86.2 & 50.5 & 37.7 & 67.6 & 36.2 & 33.5 \\ 
     (no-distractor) & EntailmentWriter (11B) & 99.0 & 89.4 & 51.5 & 38.2 & 71.2 & 38.5 & 35.3 \\
         & MetGen$^{\ast}$ & \textbf{100.0} & \textbf{100.0} & \textbf{57.7} & 41.9 & 70.8 & 39.2 & 36.5 \\
         & NLProofs$^{\ast}$ & 97.8 & 90.1 & 55.6 & \underline{42.3} & \underline{72.4} & \underline{40.6} & \textbf{38.9} \\
         & ConDec & \underline{99.9} & 98.2 & 55.7 & 42.1 & 72.3 & 38.9 & 36.2 \\
         & ConDec$^{\bigstar}$ & \underline{99.9} & \underline{98.2} & \underline{57.3} & \textbf{43.2} & \textbf{72.9} & \textbf{41.5} & \underline{37.9} \\
     \midrule
     Task 2 &  EntailmentWriter & 84.3 & 35.6 & 35.5  &  22.9 &  61.8 &  28.5 & 20.9  \\ 
     (distractor) & EntailmentWriter (11B) & 89.1 & 48.8 & 41.4 & 27.7 & 66.2 & 31.5 & 25.6 \\
         & MetGen$^{\ast}$ & 82.7 & 46.1 & 41.3 & 29.6 & 61.4 & 32.4 & 27.7 \\
         & NLProofs$^{\ast}$ & 90.3 & 58.8 & 47.2 & 34.4 & 70.2 & \underline{37.8} & 33.3 \\
         & ConDec & \underline{91.0} & \underline{59.1} & \underline{50.2} & \underline{36.5} & \underline{70.3} & \textbf{38.2} & \underline{34.1} \\
         & ConDec$^{\bigstar}$ & \textbf{91.1} & \textbf{60.6} & \textbf{50.7} & \textbf{37.4} & \textbf{70.7} & \textbf{38.2} & \textbf{34.7} \\
      \midrule
      Task 3 & EntailmentWriter & 35.7 & 2.9 & 6.1 & 2.4 & 33.4 & 7.7 & 2.4 \\ 
      (full-corpus) &  EntailmentWriter (11B) & 39.9 & 3.8 & 7.4 & 2.9 & 35.9 & 7.1 & 2.9 \\
         & MetGen$^{\ast}$ & 34.8 & \underline{8.7} & 9.8 & \textbf{8.6} & 36.6 & \textbf{20.4} & \textbf{8.6} \\
         & NLProofs$^{\ast}$ & 43.2 & 8.2 & \underline{11.2} & 6.9 & \underline{42.9} & 17.3 & 6.9 \\
         & ConDec & \underline{43.3} & 8.2 & 11.1 & 6.5 & \textbf{43.4} & \underline{18.0} & 6.5 \\
         & ConDec$^{\bigstar}$ & \textbf{44.7} & \textbf{9.4} & \textbf{11.7} & \underline{7.1} & 42.3 & 17.7 &  \underline{7.1} \\
      \bottomrule 
  \end{tabular}}
  \caption{Main results on EntailmentBank with finetuning method. Methods with $^{\ast}$ are \textbf{three-stage}. ConDec denotes finetuning with vanilla hard negatives and ConDec$^{\bigstar}$ denotes finetuning with combination of vanilla and enhanced hard negatives. Best results are \textbf{boldface} and second-best results are \underline{underlined}.}
  \vskip-0.5em
  \label{tab:ret 1}
\end{table*}

\noindent \textbf{ConDec}. The vanilla hard negatives are constructed by substituting the conclusion in the gold proof step with a randomly selected premise node context. For example, for \textit{sent11} $\&$ \textit{sent24 -> int: neptune orbits the sun in the solar system}, the hard negatives substitute conclusion \textit{neptune orbits the sun in the solar system} with the context of \textit{sent11} or \textit{sent24}. The temperature $\tau$ is set as 0.05 and $\alpha$ is 0.1. The learning rate of the stepwise decoding stage is the same as finetuning with original proof trees. The MLE loss and contrastive loss are alternatively trained for 10 epochs based on the finetuned generator with MLE loss only. \\

\noindent \textbf{Enhanced Hard Negatives}. We use Flan-T5-Large as an additional reasoner to generate unseen proof steps based on labeled proof trees as hard negatives. To train the reasoner, details of data collection are presented in Appendix \ref{sec:appendix 1}. 
The reasoner is trained for 30 epochs with a learning rate of 1e-4.
We further apply Vera \cite{liu2023vera} to filter unreasonable generated proof steps. Details of stepwise proofs sampled for reasoner and enhanced new negatives filtered by the checker are presented in Table \ref{tab:data 1}. To increase diversity, enhanced hard negatives and vanilla hard negatives are jointly sampled for training.



\section{Result Analysis}

\subsection{Main Results}
The main results are presented in Table \ref{tab:ret 1}. By analyzing the results, we can find that: 

\noindent
\textbf{Stepwise correction matters}. Stepwise generation methods (MetGen, NLProofs, ConDec) outperform single-shot training (EntailmentWriter), even for EntailmentWriter with a much larger parameter size (11B). When comparing ConDec with three-stage generation methods MetGen and NLProofs, ConDec achieves comparable or even better performance on Task 2 and Task 3. This shows that contrastive decoding with hard negatives can improve language models' reasoning ability, demonstrating our methods' effectiveness. While our research focuses on the generator, combining our method with theirs may still improve the final accuracy and it is worth exploring.

\noindent
\textbf{Enhanced hard negatives facilitate reasoning}. With enhanced hard negatives, we can find that the ability of proof planning over three tasks is all improved. Unlike vanilla negative construction, the enhanced hard negatives contain harder or more accurate conclusions given premises. Detailed evaluation of enhanced hard negatives is in Appendix \ref{sec:appendix 1}. It further improves the training coverage over reasoning steps, thus leading to better performance over proof planning.

\noindent
\textbf{ConDec is more efficient with distractors}. Specifically, contrastive learning with hard negatives achieves obvious performance gain over Task 2 and Task 3 while deteriorating the performance on Task 1. For Task 2 and Task 3, there are distractor premises or full-corpus, which increases the challenges for rigorous reasoning. For Task 1, there is no distractor. Adding hard negatives deviates the generator from generating correct steps instead of intermediate nodes.

\noindent
\textbf{Predicting intermediate node is still challenging}. ConDec achieves the best performance mostly over leave or step accuracy. The contrastive loss helps the generator discriminate between premises and finds semantically correlated premises to deduce a conclusion. However, deductive reasoning is still challenging as the improvement over intermediates is not as obvious as that on leaves or steps. 

\subsection{Computational Cost Analysis}
ConDec's design can reduce the inference cost with a one-stage process during inference as shown in Table \ref{tab:computation}. We provide a concrete comparison using the validation set for Task 2, with the same model size (Flan-T5-Large), on an A800 GPU. It clearly shows that ConDec can significantly improve time efficiency compared to two and three-stage methods (NLProofs).

\subsection{Ablation Study}
Results of the ablation study on Task 2 test split are presented in Table \ref{tab:ret 3}. 

\begin{table}[]
\resizebox{0.5\textwidth}{!}{
\begin{tabular}{llcc}
\toprule
Method   & Stage    & Time(s) & Improvement \\
\midrule
ConDec   & generator    & 294   & -      \\
NLProofs & generator+search   & 430  & 31.6\%      \\
NLProofs & generator+search+verify & 544  & 46.0\% \\
\bottomrule
\end{tabular}
}
\caption{Computation time comparison(percentage reduction). Time denotes the inference time. NLProofs requires additional search and verification for prediction.}
\label{tab:computation}
\vspace{-1.0em}
\end{table}


\begin{table*}[t]
  \centering
  \resizebox{\textwidth}{!}{
  \begin{tabular}{lccccccc}
  \toprule
  \multicolumn{1}{c}{\multirow{2}{*}{Method}} & \multicolumn{2}{c}{Leaves}                              & \multicolumn{2}{c}{Steps}                               & \multicolumn{2}{c}{Intermediates}                       & \multicolumn{1}{c}{Overall}    \\
  \multicolumn{1}{c}{}   & \multicolumn{1}{c}{F1} & \multicolumn{1}{c}{AllCorrect} & \multicolumn{1}{c}{F1} & \multicolumn{1}{c}{AllCorrect} & \multicolumn{1}{c}{F1} & \multicolumn{1}{c}{AllCorrect} & \multicolumn{1}{c}{AllCorrect} \\
    \midrule
    LLaMA-3.2-1B (1.2B, w/o stepwise) & 19.5 & 5.3 & 6.4 & 2.9 & 13.9 & 5.3 & 2.7 \\
    LLaMA-3.2-1B (1.2B, stepwise) & 15.6 & 3.5 & 4.1 & 3.2 & 9.9 & 5.0 & 3.2 \\
    \midrule
    Flan-T5-Large (0.8B, stepwise) & 90.7 & 58.8 & 49.2 & 36.2 & 69.6 & 36.8 & 33.5 \\
    + contrastive loss & 90.9 & 60.3 & 49.5 & 35.9 & 69.4 & 37.1 & 32.4 \\
    + contrastive loss, vanilla hard & 91.0 & 59.1 & 50.2 & 36.5 & 70.3 & \textbf{38.2} & 34.1 \\
    + contrastive loss, vanilla and enhanced hard & \textbf{91.1} & \textbf{60.6} & \textbf{50.7} & \textbf{37.4} & \textbf{70.7} & \textbf{38.2} & \textbf{34.7} \\
    \midrule
    Flan-T5-XL (3B, stepwise) & 90.9 & 57.1 & 50.2 & 36.5 & 68.8 & 35.9 & 33.8 \\
    \bottomrule 
  \end{tabular}}
  \caption{Ablation study results on EntailmentBank test set in Task 2. }
  \label{tab:ret 3}
  \vskip-0.5em
\end{table*}
\begin{table*}[t]
  \centering
  \small
  \begin{tabular}{p{3.2cm}p{1.25cm}p{1.25cm}p{1.25cm}p{1.25cm}p{1.25cm}p{1.25cm}p{1.25cm}}
  \toprule
  \multicolumn{1}{c}{\multirow{2}{*}{Method}} & \multicolumn{2}{c}{Leaves}                              & \multicolumn{2}{c}{Steps}                               & \multicolumn{2}{c}{Intermediates}                       & \multicolumn{1}{c}{Overall}    \\
  \multicolumn{1}{c}{}   & \multicolumn{1}{l}{F1} & \multicolumn{1}{c}{AllCorrect} & \multicolumn{1}{l}{F1} & \multicolumn{1}{c}{AllCorrect} & \multicolumn{1}{l}{F1} & \multicolumn{1}{c}{AllCorrect} & \multicolumn{1}{c}{AllCorrect} \\
    \midrule
    ConDec$^{\bigstar}$ & \textbf{92.4} & \textbf{63.1} & \textbf{55.3} & \textbf{45.5} & \textbf{72.8} & \textbf{43.9} & \textbf{41.2} \\
    \midrule
    GPT3 ($5$-shot) & 64.2 & 15.3 & 17.6 & 12.3 & 53.6 & 22.3 & 12.3 \\
    GPT3.5-turbo ($5$-shot) & 61.9 & 9.0 & 16.9 & 4.3 & 51.9 & 15.1 & 3.74 \\
    GPT4 ($5$-shot) & 78.1 & 32.6 & 30.2 & 22.5 & 63.9 & 30.8 & 21.9  \\
    GPT4 (SI) & 79.1 & 30.0 & 24.3 & 14.0 & 65.4 & 32.1 & 13.9 \\ 
    GPT4 (CoT) & 79.0 & 33.7 & 30.9 & 22.5 & 64.7 & 31.6 & 22.5 \\   
    \midrule
    GPT4o-mini ($5$-shot) & 63.1 & 18.2 & 20.0 & 13.9 & 54.5 & 24.1 & 13.9 \\
    o1-mini ($5$-shot) & 72.5 & 27.3 & 28.6 & 13.9 & 50.9 & 18.7 & 11.8 \\
    o1-preview ($5$-shot) & 84.0 & 43.3 & 38.7 & 28.0 & 67.8 & 33.7 & 21.9 \\
    \bottomrule 
  \end{tabular}
  \caption{Results on EntailmentBank validation set in Task 2 with finetuning and prompting methods. GPT3.5 is \textit{GPT3.5-turbo-0613}. GPT4 is \textit{GPT4-0613}. GPT4o-mini is \textit{gpt-4o-mini-2024-07-18}. o1-mini is \textit{o1-mini-2024-09-12}. o1-preview is \textit{o1-preview-2024-09-12}.}
  \label{tab:ret 2}
  \vskip-0.5em
\end{table*}

\noindent \textbf{Backbone model}. The results indicate that LLaMA-3.2-1B fails to adequately differentiate between input premises and output reasoning proof steps. As a decoder model focused on next-word prediction, LLaMA simply predicts proof steps as the next sentence following the premises. With stepwise training, the presence of intermediate or hypothesis nodes can confuse the model regarding when to stop generating output, as a subtree with an intermediate node is a subsequence of the entire tree that includes the hypothesis node. This is why LLaMA tends to produce shorter outputs when trained in a stepwise manner.

Moreover, the results for LLaMA are significantly lower than those for Flan-T5-Large, despite the latter's smaller size. Natural proof generation is fundamentally a sequence-to-sequence task. Distinguishing between input and output is crucial, as the encoder-decoder architecture of Flan-T5 processes different texts. A case study comparing LLaMA and Flan-T5 is presented in the Appendix \ref{append:case}.\\
\noindent \textbf{Model size}. For the model size, we also try Flan-T5-XL (3B). Stepwise training over Flan-T5-Large provides a strong baseline. Though much larger in model size, Flan-T5-XL achieves similar performance over the metrics. It shows that proof planning requires a higher level of understanding of deductive reasoning over multi-hops. \\
\noindent \textbf{Negative sample}. ConDec without hard negatives improves the leave accuracy while decreasing the step accuracy, leading to a slight drop in the overall performance. Adding vanilla or enhanced hard negatives can generally improve over all the metrics. 

\section{Analysis of Closed-Source LLMs}
In-context learning (ICL; \citealt{brown2020language}) and CoT~\citep{WeiCoT22} have been widely used in various reasoning tasks \cite{patel2021nlp, cobbe2021training, fu2022complexity, Xiong2023DQLoReDQ}. We apply 5-shot prompting and CoT to evaluate how LLMs perform on the proof generation task. The vanilla prompt method simply adopts a $k$-shot in-context learning strategy. The CoT decomposes the reasoning process into step-by-step generation. Based on it, Select-inference(SI)\cite{creswell2022selection} is a two-stage COT that decomposes each reasoning step into a premise selection stage and a conclusion inference state. Details of the CoT is illustrated in Appendix~\ref{sec:appendix 3-1}. 

As shown in Table \ref{tab:ret 2}. Comparing different LLMs, GPT4 generally outperforms GPT3 and GPT3.5-turbo on all the metrics.  
One detailed case study of GPT3.5-turbo and GPT4 with $k$-shot prompting results is in Appendix \ref{sec:appendix 3-2}. It shows that the GPT3.5-turbo tends to generate more irrelevant and inaccurate steps with simple imitation of premises during inferencing new conclusions. In contrast, GPT4 conducts deductive reasoning and generates correct conclusions based on premises. When accompanied by CoT, GPT4 achieves better performance than the vanilla prompt. While with SI, GPT4 makes more mistakes in the premise selection stage. 

Our ConDec$^{\bigstar}$  outperforms all closed-source LLMs in all metrics. Although recent LLMs such as o1-mini and o1-preview show improvements in several metrics compared to their predecessors (GPT-3.5-turbo and GPT-4), a substantial performance gap remains when compared to finetuning-based methods. Through stepwise training with curated datasets, these language models can better capture the correlations between premises.

\section{Conclusion}
Logical reasoning is both challenging and fundamental in artificial intelligence. Proof generation serves as a measure of the explanatory capabilities of large language models in the context of logical reasoning. To enhance stepwise deductive reasoning, we propose a decoding strategy augmented by contrastive learning. The carefully designed hard negatives address the typical errors encountered during the decoding process. The experimental results across standard benchmarks demonstrate the effectiveness of our method. Additionally, our analysis of larger LLMs reveals that they continue to struggle with proof planning tasks.

\section*{Limitations}
From the analysis of the paper, natural proof planning ability is still a challenging topic in evaluating LLMs' deductive reasoning ability. The current curated human-annotated dataset in our experiments is of limited size to improve LLMs' deductive reasoning ability. Knowledge from related corpora such as cause and effect, logic reasoning can be further applied to improve the proof generation ability with pre-training or transfer learning.

\bibliography{anthology,custom}
\bibliographystyle{acl_natbib}

\clearpage

\appendix

\section{Appendix}
\label{sec:appendix}

\subsection{Details of Enhanced Hard Negatives}
\label{sec:appendix 1}
Training data of the reasoner is basically all the proof steps in the training set. Each proof step is converted into the form of natural language: 1) premises are concatenated with conjunction ``and''; 2) premises and conclusion are linked with ``Because'' and ``Therefore''. The reasoner is trained with 8,819 proof steps for both Task 1 and Task 2.  
 
For each proof step ${\rm {step}_j}$, substitute one premise from the the supporting fact set $C$ except the ones in current proof step. Recombined proof premises $\{p^{s}_1, p^{s}_2, ...\}$ are converted into natural language form and used as input to the reasoner. The reasoner generates a new conclusion $c^{s}$ based on the premises, forming a hard negative ${\overline{\rm {step}}}^{(i)}_j$. 

The constructed hard negative ${\overline{\rm {step}}}^{(i)}_j$ is then filtered by the checker with a threshold score. To determine the threshold score, three PhD students from the CSE department are assigned with the task to check the quality of 100 randomly selected constructed hard negatives. Each hard negative is first judged by Vera. The $score$ from Vera indicates the reasonableness of the hard negative. By filtering the $score$ with a threshold, we choose the constructed negatives with high quality for further training. The result of human checking over the filtered negatives is shown in the table \ref{tab:append 1}. We find that with a threshold score of 0.9, 60\% of the filtered hard negatives are reasonable. 

\begin{table}[]
  \centering
  \begin{tabular}
  {cc}
  \toprule
  Threshold score & Accuracy(\%) \\
  \midrule
  0.7 & 36 \\
  0.8 & 50 \\
  0.9 & 60 \\
  \bottomrule
  \end{tabular}
  \caption{Human check accuracy on enhanced hard negatives from reasoner and checker with different threshold scores.}
  \label{tab:append 1}
\end{table}

\subsection{Prompting Methods}
\subsubsection{Case study for prompting}
\label{sec:appendix 3-2}
Result case from GPT3.5-turbo and GPT4 on Task2 dev split with vanilla prompts is shown in Table \ref{tab:append 3}. 

\subsubsection{Template for prompting}
\label{sec:appendix 3-1}
Prompting template for GPT4 with CoT is in Table \ref{tab:append 4}. 

\subsection{Baseline Details}
MetGen and NLProofs are three-stage methods. MetGen divides single-step entailment into basic logical operations. It reasons in both forward deductive and backward abductive steps. A controller finally selects promising steps among the reasoning results. NLProofs uses a trained verifier to guide the search process of proof generation. The verifier scores the generation expansion steps and finally, a proof tree is selected according to the scores. \\

\subsection{Discussion with RLET}

RLET\cite{liu2022rlet} is trained using cumulative signals across the entire entailment tree, marking the first introduction of reinforcement learning into the entailment tree generation task. It employs a one-stage generation method that does not involve search or verification. RLET flexibly assigns rewards to each generated step and utilizes the overall cumulative reward to optimize training based on the full trajectory. In contrast, our approach and previous methods (as listed in Table \ref{tab:1}) rely on exact matches of gold steps as training signals.

In Task 3, the retrieved supporting facts may contain noise and are longer than those in Task 1 and Task 2. RLET’s focus on long-path proof training provides it with an advantage over our method (and previous methods listed in Table 1) in this task. However, since the supporting facts are sourced from a relevant corpus, they may deviate from the training setting, particularly under zero-shot performance conditions. As a result, RLET's performance is significantly lower than that of previous methods in Task 1 and Task 2.



\begin{table*}[]
    \centering
    \begin{tabular}{p{16cm}}
    \toprule
    \textbf{Hypothesis}: new york state has the least sunlight during december \\
    \textbf{Context}: \\
    sent1: precipitation is a kind of weather \\
    sent2: daylight hours means time during which there is daylight \\
    sent3: state means form \\
    sent4: water is in the liquid state , called liquid water , for temperatures between 273 / 32 / 0 and 373 / 212 / 100 k / f / c \\
    sent5: united states is located in the northern hemisphere \\
    ... \\
    sent9: when the seasons change from summer to fall / from summer to winter , the amount of daylight will decrease \\
    ... \\
    sent14: new york / new york state is a state located in the united states of america \\
    sent15: winter has the least sunlight \\
    ... \\
    sent22: the earth being tilted on its rotating axis causes seasons \\
    sent23: december is during the winter in the northern hemisphere \\
    ...\\
    \midrule
     \textbf{Gold Proof}: sent14 \& sent5 -> int1: new york state is located in the northern hemisphere; int1 \& sent23 -> int2: december is during the winter for new york state; int2 \& sent15 -> hypothesis; \\
    \midrule
    \textbf{GPT3.5-turbo}: sent23 \& sent22 -> int1: December is during the winter in the northern hemisphere; \\
    int1 \& sent9 -> int2: the amount of daylight will decrease during December; int2 \& sent15 -> int3: December has the least sunlight; int3 \& sent14 -> hypothesis; \\
    \midrule
    \textbf{GPT4}: sent14 \& sent5 -> int1: New York state is located in the northern hemisphere; int1 \& sent23 -> int2: December is during the winter in New York state; int2 \& sent15 -> hypothesis; \\
    \bottomrule
    \end{tabular}
    \caption{Case study of vanilla prompt with GPT3.5-turbo and GPT4.}
    \label{tab:append 3}
\end{table*}

\begin{table*}[]
    \centering
    \begin{tabular}{p{16cm}}
    \toprule
    \textbf{Hypothesis}: if a fossil of a bird cannot be identified then that kind of bird is probably extinct. \\
    \textbf{Context}: \\
     sent1: identifying is similar to determining. \\
     sent2: if a fossil is of an organism that cannot be identified then that organism is probably extinct. \\
     sent3: discovering something usually requires seeing that something. \\
     sent4: a dinosaur is a kind of extinct animal. \\
     sent5: fossils can be used as evidence for the ancient environment. \\
     sent6: dead means not alive.\\
     ...\\
     sent25: fossils can be used to study the history of organisms and environments on earth \\
     \\
     Reasoning step by step and finally output the proof: \\
     From sent13 \& sent24, we can infer that a bird is a kind of organism (int1); \\
     From int1 \& sent2, we can infer that if a fossil of a bird cannot be identified then that kind of bird is probably extinct (hypothesis); \\
     Proof: sent13 \& sent24 -> int1: a bird is a kind of organism; int1 \& sent2 -> hypothesis; \\
     \bottomrule
    \end{tabular}
    \caption{Prompting template of stepwise proof generation for GPT4 with COT.}
    \label{tab:append 4}
\end{table*}

\newpage
\subsection{Selection of Enhanced Hard Negatives}
Besides randomly selecting premises to construct enhanced hard negatives, we also select according to the top similarity score calculated with BM25 \cite{robertson2009probabilistic}. Results are presented in Table \ref{tab:append 2}. The results show that similar distracted premises do not contribute to intermediate node accuracy as much as random premises do. This is because randomly selected premises can introduce more diversity for the constructed hard negatives.

\begin{table*}[]
  \centering
  \scalebox{0.8}{
  \begin{tabular}{lccccccc}
  \toprule
  \multicolumn{1}{c}{\multirow{2}{*}{Method}} & \multicolumn{2}{c}{Leaves}                              & \multicolumn{2}{c}{Steps}                               & \multicolumn{2}{c}{Intermediates}                       & \multicolumn{1}{c}{Overall}    \\
  \multicolumn{1}{c}{}   & \multicolumn{1}{c}{F1} & \multicolumn{1}{c}{AllCorrect} & \multicolumn{1}{c}{F1} & \multicolumn{1}{c}{AllCorrect} & \multicolumn{1}{c}{F1} & \multicolumn{1}{c}{AllCorrect} & \multicolumn{1}{c}{AllCorrect} \\
    \midrule
    Random & 91.0 & 60.6 & 50.7 & 37.4 & 70.7 & 38.2 & 34.7 \\
    \midrule
    BM25 & 90.8 & 60.3 & 50.3 & 36.8 & 68.8 & 37.9 & 34.7 \\
    \bottomrule 
  \end{tabular}}
  \caption{Ablation study of enhanced hard negative construction on EntailmentBank test set on Task 2. }
  \label{tab:append 2}
\end{table*}

\subsection{Case Study}
\label{append:case}
We present a case study comparing the generation capabilities of LLaMA 3.2-1B and Flan-T5-Large to illustrate the differences between encoder-decoder and decoder-only LLMs, as shown in Table \ref{tab:append 5}. The table indicates that LLaMA (without stepwise training) generates longer proofs because it is trained with complete proof chains. In contrast,LLaMA (with stepwise training) typically produces one-step proofs that conclude with either a "hypothesis" or an "intermediate node," reflecting the fact that both are endpoints in the sequences used for stepwise training. Flan-T5-Large (with stepwise training) strikes a balance between generating long and short proof chains due to its encoder-decoder architecture. However, it still faces challenges in accurately selecting premises.

\begin{table*}[]
    \centering
    \begin{tabular}{p{16cm}}
    \toprule
    \textbf{Hypothesis}: as the distance of the star to earth decreases, the star will appear brighter. \\
    \textbf{Context}: \\
     sent1: a star produces light and heat \\
     sent2: far is the opposite of close \\
     sent3: as the distance from an object increases , the force of gravity on that object will decrease \\
     sent4: brightness means amount of light \\
     sent5: if two or more things are in a relationship then those things impact each other \\
     sent6: feature is synonymous with characteristic\\
     ...\\
     sent25: moving away from the source increases the distance \\
     \\
     \midrule
     \textbf{Gold Proof}: sent1 \& sent13 -> int1: a star is a source of light; int1 \& sent21 \& sent23 -> hypothesis; \\
     \midrule
     \textbf{LLaMA-3.1-1B(w/o stepwise)}: sent13 \& sent1 -> int1: stars are a source of light; int1 \& sent21 -> int2: as the distance of a star to earth decreases, the star will appear brighter; int2 \& sent7 -> hypothesis; \\
     \midrule
     \textbf{LLaMA-3.1-1B(stepwise)}: sent21 \& sent3 -> hypothesis; \\
     \midrule
     \textbf{Flan-T5-Large(stepwise)}: sent1 \& sent13 -> int1: a star is a source of light and produces light; sent17 \& int1 \& sent21 -> hypothesis;\\
     \bottomrule
    \end{tabular}
    \caption{Case study of LLaMA and Flan-T5.}
    \label{tab:append 5}
\end{table*}



\end{document}